\documentclass[journal]{IEEEtran}
\pdfoutput=1
\usepackage{cite}
\usepackage{amsmath,amssymb,amsfonts}
\usepackage{graphicx}
\usepackage{subfigure}
\usepackage{booktabs}
\usepackage{multirow}
\usepackage{booktabs}
\usepackage{threeparttable}
\usepackage{underscore}
\usepackage{hyperref}
\usepackage{color}
 
\begin{document}
 
\title{Universal Deep Network for Steganalysis of Color Image based on Channel Representation}

\author{Kangkang Wei,
        Weiqi Luo*, ~\IEEEmembership{Senior Member,~IEEE,  }
        Shunquan Tan, ~\IEEEmembership{Senior Member,~IEEE, } 
        Jiwu Huang, ~\IEEEmembership{Fellow,~IEEE}

\thanks{K. Wei  and W. Luo are with the Guangdong Key Lab of Information Security Technology and School of Computer Science and Engineering, Sun Yat-sen University, Guangzhou 510006, China (e-mail: weikk5@mail2.sysu.edu.cn, luoweiqi@mail.sysu.edu.cn).}
\thanks{S. Tan  is with the College of Computer Science and Software Engineering, Shenzhen University, Shenzhen 518060, China (e-mail: tansq@szu.edu.cn). } 
\thanks{J. Huang is with the Guangdong Key Laboratory of Intelligent Information Processing, Shenzhen Key Laboratory of Media Security, and National Engineering Laboratory for Big Data System Computing Technology, Shenzhen University, Shenzhen 518060, China (e-mail: jwhuang@szu.edu.cn).} 
\thanks{*W. Luo is the corresponding author.}
}
  \maketitle

\begin{abstract}
Up to now,  most existing steganalytic methods are  designed for grayscale images, and they are not  suitable for color images that are widely used in current social networks. In this paper,  we design a universal color image steganalysis network (called UCNet)  in  spatial  and JPEG domains.  The proposed method includes preprocessing, convolutional, and classification modules.  To preserve the steganographic artifacts in each color channel,  in preprocessing module, we firstly separate the input image into three channels according to the corresponding embedding spaces (i.e. RGB for spatial steganography and YCbCr for JPEG steganography), and then extract the  image residuals with 62 fixed  high-pass filters, finally concatenate all truncated residuals for subsequent analysis  rather than adding them together with normal convolution like existing CNN-based steganalyzers.   To accelerate the network convergence and effectively reduce the number of parameters, in convolutional module, we carefully design three types of layers with different shortcut connections and group convolution structures to  further learn high-level steganalytic features.  In classification module, we employ a global average pooling and fully connected layer for classification. We conduct extensive experiments on ALASKA II to demonstrate that the proposed method can achieve state-of-the-art results compared with the modern CNN-based steganalyzers (e.g.,  SRNet and J-YeNet) in both spatial and JPEG domains, while keeping relatively few memory requirements and training time.  Furthermore, we also provide  necessary  descriptions and many ablation experiments to verify the rationality of the network design.
\end{abstract}


\IEEEpeerreviewmaketitle

\section{Introduction}

\IEEEPARstart{I}{mage} steganography aims to embed secret information into digital image without introducing obvious visual artifacts. On the opposite side, steganalysis aims to detect covert communication established via steganography. Recently, most modern steganographic methods (\cite{luo2010edg,fil2010gib,lib2014ane,sed2016con,guo2015usi}) are image content adaptive, which significantly enhance the steganography security. Thus, steganalysis is facing severe challenges. 

Image steganalysis techniques can be divided into two categories, that is, traditional methods based on handcrafted features and CNN-based methods.  Feature extraction is the key issue in traditional methods. For instance, in spatial domain, Penvy \textit{et al.} \cite{pev2010ste} obtained 686-dimensional features by extracting inter-pixel relationships. Fridrich \textit{et al.} \cite{fri2012ric} used several different high-pass filters and calculated the co-occurrence matrix for the obtained features to get 34671-dimensional features (called SRM). Holub \textit{et al.} \cite{hol2013ran} proposed a random projection feature analysis method based on the residuals of SRM. In JPEG domain,  Kodovský \textit{et al.} \cite{jab2012ste} presented a rich model of JPEG to capture the changes brought by the steganographic signal more comprehensively. Holub \textit{et al.} \cite{hol2015low} proposed a DCTR feature with 64 kernels using discrete cosine transform. Later, Song \textit{et al.} \cite{song2015ste} used Gabor filters instead of SRM kernels to detect JPEG steganography, and also achieved better performance. Recently, many CNN-based steganalytic methods (e.g., \cite{qia2015dee,xu2017dee,ye2017dee,che2017jpe,yed2018yed,bor2019dee,hua2019acu,den2019fas,zha2020dep}) have been proposed  and achieve much better results compared with traditional ones. For instance, in spatial domain, Ye \textit{et al.} \cite{ye2017dee} proposed a CNN structure and also with the help of selection channel knowledge, the model gained better detection accuracy over the SRM (called YeNet). Yedroudj \textit{et al.} \cite{yed2018yed} achieved further performance improvement based on YeNet using data augmentation, batch normalization (called Yedroudj-Net). Deng \textit{et al.} \cite{den2019fas} introduced covariance pooling to the steganalysis task (we call it CovNet) and achieved better results. In JPEG domain, Xu \cite{xu2017dee} designed a 20-layer CNN-based steganalyzer called J-XuNet for detecting J-UNIWARD \cite{hol2014uni}. Boroumand \textit{et al.} \cite{bor2019dee} described a universal deep steganalysis residual network (called SRNet), and showed its effectiveness in both spatial and JPEG steganography. Huang \textit{et al.} \cite{hua2019acu} presented a  selection-channel-aware CNN called J-YeNet by extending the YeNet \cite{ye2017dee} for JPEG steganalysis.

Note that  all above mentioned methods are  designed for detecting grayscale stegos, and they are not very suitable for detecting color stegos generated by some steganography methods, such as  \cite{tan2015clu,qin2019ano,wan2020non,tab2018ane, cog2019the,cog2020ste}.  Recently,  several  traditional steganalytic methods,  such as \cite{lia2016con,abd2016col,gol2014ric, yan2020col},  have been proposed for color images.  For instance, Goljan  \textit{et al.} \cite{gol2014ric} proposed an extension of the SRM \cite{fri2012ric} for steganalysis of color images (called CRM).  Yang \textit{et al.}  \cite{yan2020col} presented a variant of SRM feature based on the embedding change probabilities in differential channels between different color channels.  In spatial domain, Zeng \textit{et al.} \cite{zen2019wis} first design a wider separate-then-reunion network (called WISERNet) for steganalysis of color images,  and achieve better results  compared with related methods based on handcrafted features.   Except for WISERNet,  to our best knowledge,  the recent CNN-based steganalyzers for color images are directly based on some existing methods for grayscale images or some effective models in computer vision.  For instance, in JPEG domain,  Yousfi \textit{et al.} \cite{you2020ima} and Chubachi \cite{chu2020ane} adjust the existing networks (i.e., SRNet)  for color JPEG images.   Furthermore, they introduce other improvement strategies, such as model pre-training on a big image dataset (e.g., ImageNet \cite{den2009ima}) or ensemble model with different CNNs, to further enhance detection performances of the existing networks.   Butora and Yousfi \textit{et al.}  \cite{but2021how,you2021imp} explored the effect of several variants of EfficientNet in steganalysis. Although these steganalyzers can achieve satisfactory results,  some inherent steganography artifacts within color images have  not been fully exploited for steganalysis, such as  the relationship between color channels.  Like SRNet for grayscale images,  in addition,  a universal steganalytic network  that is effective in both spatial and JPEG domains for color images is critically needed. 
 
Aiming at color images,  we design a universal deep steganalytic network based on channel representation in this paper. In the proposed method,  we firstly separate the color input image into three channels according to their embedding spaces (i.e. RGB for spatial steganography and YCbCr for JPEG steganography). To enhance the steganographic noise signal  in both spatial and JPEG domains, we combine 30 basic linear filters from SRM and 32 Gabor filters for calculating image residuals.  To well preserve steganographic artifacts in each color channel, we then concatenate all truncated residuals for subsequent analysis  instead of adding them together like existing steganalytic networks.  Furthermore,  we carefully design three types of layers with  different shortcut connections and group convolution structures to further learn high-level steganalytic features.  Extensive experimental results evaluated on ALASKA II show that  the proposed method can  achieve state-of-the-art results compared with some  modern CNN-based steganalyzers,  while maintaining lower resource requirements and number of parameters.  In addition, sufficient ablation experiments are also provided to verify the rationality of the network design. 
 
The rest of the paper is organized as follows.  Section \ref{Sec:Proposed} describes the proposed method in detail. Section \ref{Sec:Experiments} shows comparative experimental results and discussions.  Finally, the concluding remarks of this paper and future works are given in Section \ref{Sec:Conclusion}.

\section{Conclusion remarks}
\label{Sec:Conclusion}
Most existing steganalytic methods are originally designed for detecting grayscale images, and they are not very effective in color image steganalysis based on our experiments.  In this paper, therefore,  we proposed a universal deep network for steganalysis of color image based on  channel representation, and demonstrate that  the proposed method can achieve the best detection performances for compared with some modern steganalytic networks  {in both} spatial and JPEG domains. The major contributions of this work are as follows:

\begin{itemize}
\item  To well preserve the steganographic artifacts in color images,  we employ the color channel separation and then concatenate feature maps instead of  the convolution summation which is commonly used in existing steganalytic networks; 
\item To extract the high-level steganalytic features, we carefully design  three new types of layers and  then combined them into a continuous convolutional module.  Experimental results on ALASKA II  demonstrate the superiority of the proposed steganalyzer compared with the modern steganalytic networks  {in both} spatial and JPEG domains;  
\item  To verify the rationality of our proposed model,  we give some necessary descriptions about our network design in three modules (i.e., preprocessing, convolutional, and classification). Furthermore, we provide extensive ablation experiments of the proposed   model. 
\end{itemize}

Our future work will focus on two aspects: 1) we will integrate {selection-channel-aware} into the proposed steganalyzer, and adaptively enhance the model features from different levels of network.  In addition,  we will design an effective strategy to improve inter-channel correlation by combining embedding probability information for each channel of color images; 2) We will further explore deep learning-based architectures for color image steganalysis using inter-channel correlation, such as introducing quaternion convolutional neural networks to preserve inter-channel dependencies and thus extract richer steganalytic features.


\ifCLASSOPTIONcaptionsoff
  \newpage
\fi


\bibliographystyle{ieeetr}
\bibliography{cited}{}

\begin{thebibliography}{10}

\bibitem{luo2010edg}
W.~Luo, F.~Huang, and J.~Huang, ``Edge adaptive image steganography based on
  lsb matching revisited,'' {\em IEEE Transactions on Information Forensics and
  Security}, vol.~5, no.~2, pp.~201--214, 2010.

\bibitem{fil2010gib}
T.~Filler and J.~Fridrich, ``Gibbs construction in steganography,'' {\em IEEE
  Transactions on Information Forensics and Security}, vol.~5, no.~4,
  pp.~705--720, 2010.

\bibitem{lib2014ane}
B.~Li, M.~Wang, J.~Huang, and X.~Li, ``A new cost function for spatial image
  steganography,'' in {\em IEEE International Conference on Image Processing},
  pp.~4206--4210, 2014.

\bibitem{sed2016con}
V.~Sedighi, R.~Cogranne, and J.~Fridrich, ``Content-adaptive steganography by
  minimizing statistical detectability,'' {\em IEEE Transactions on Information
  Forensics and Security}, vol.~11, no.~2, pp.~221--234, 2016.

\bibitem{guo2015usi}
L.~Guo, J.~Ni, W.~Su, C.~Tang, and Y.-Q. Shi, ``Using statistical image model
  for \protect{JPEG} steganography: Uniform embedding revisited,'' {\em IEEE
  Transactions on Information Forensics and Security}, vol.~10, no.~12,
  pp.~2669--2680, 2015.

\bibitem{pev2010ste}
T.~Pevny, P.~Bas, and J.~Fridrich, ``Steganalysis by subtractive pixel
  adjacency matrix,'' {\em IEEE Transactions on Information Forensics and
  Security}, vol.~5, no.~2, pp.~215--224, 2010.

\bibitem{fri2012ric}
J.~Fridrich and J.~Kodovsky, ``Rich models for steganalysis of digital
  images,'' {\em IEEE Transactions on Information Forensics and Security},
  vol.~7, no.~3, pp.~868--882, 2012.

\bibitem{hol2013ran}
V.~Holub and J.~Fridrich, ``Random projections of residuals for digital image
  steganalysis,'' {\em IEEE Transactions on Information Forensics and
  Security}, vol.~8, no.~12, pp.~1996--2006, 2013.

\bibitem{jab2012ste}
J.~Kodovský and J.~Fridrich, ``{Steganalysis of \protect{JPEG} images using
  rich models},'' in {\em Media Watermarking, Security, and Forensics},
  vol.~8303, pp.~81--93, 2012.

\bibitem{hol2015low}
V.~Holub and J.~Fridrich, ``Low-complexity features for \protect{JPEG}
  steganalysis using undecimated dct,'' {\em IEEE Transactions on Information
  Forensics and Security}, vol.~10, no.~2, pp.~219--228, 2015.

\bibitem{song2015ste}
X.~Song, F.~Liu, C.~Yang, X.~Luo, and Y.~Zhang, ``Steganalysis of adaptive
  \protect{JPEG} steganography using 2\protect{D} gabor filters,'' in {\em ACM
  Workshop on Information Hiding and Multimedia Security}, pp.~15--–23, 2015.

\bibitem{qia2015dee}
Y.~Qian, J.~Dong, W.~Wang, and T.~Tan, ``Deep learning for steganalysis via
  convolutional neural networks,'' in {\em Media Watermarking, Security, and
  Forensics}, vol.~9409, p.~94090J, 2015.

\bibitem{xu2017dee}
G.~Xu, ``Deep convolutional neural network to detect j-uniward,'' in {\em ACM
  Workshop on Information Hiding and Multimedia Security}, pp.~67–--73, 2017.

\bibitem{ye2017dee}
J.~Ye, J.~Ni, and Y.~Yi, ``Deep learning hierarchical representations for image
  steganalysis,'' {\em IEEE Transactions on Information Forensics and
  Security}, vol.~12, no.~11, pp.~2545--2557, 2017.

\bibitem{che2017jpe}
M.~Chen, V.~Sedighi, M.~Boroumand, and J.~Fridrich,
  ``\protect{JPEG}-phase-aware convolutional neural network for steganalysis of
  \protect{JPEG} images,'' in {\em ACM Workshop on Information Hiding and
  Multimedia Security}, pp.~75--84, 2017.

\bibitem{yed2018yed}
M.~Yedroudj, F.~Comby, and M.~Chaumont, ``Yedroudj-\protect{N}et: An efficient
  \protect{CNN} for spatial steganalysis,'' in {\em IEEE International
  Conference on Acoustics, Speech and Signal Processing}, pp.~2092--2096, 2018.

\bibitem{bor2019dee}
M.~Boroumand, M.~Chen, and J.~Fridrich, ``Deep residual network for
  steganalysis of digital images,'' {\em IEEE Transactions on Information
  Forensics and Security}, vol.~14, no.~5, pp.~1181--1193, 2019.

\bibitem{hua2019acu}
J.~Huang, J.~Ni, L.~Wan, and J.~Yan, ``A customized convolutional neural
  network with low model complexity for \protect{JPEG} steganalysis,'' in {\em
  ACM Workshop on Information Hiding and Multimedia Security}, pp.~198–--203,
  2019.

\bibitem{den2019fas}
X.~Deng, B.~Chen, W.~Luo, and D.~Luo, ``Fast and effective global covariance
  pooling network for image steganalysis,'' in {\em ACM Workshop on Information
  Hiding and Multimedia Security}, pp.~230–--234, 2019.

\bibitem{zha2020dep}
R.~Zhang, F.~Zhu, J.~Liu, and G.~Liu, ``Depth-wise separable convolutions and
  multi-level pooling for an efficient spatial \protect{CNN}-based
  steganalysis,'' {\em IEEE Transactions on Information Forensics and
  Security}, vol.~15, pp.~1138--1150, 2020.

\bibitem{hol2014uni}
V.~Holub, J.~Fridrich, and T.~Denemark, ``Universal distortion function for
  steganography in an arbitrary domain,'' {\em EURASIP Journal on Information
  Security}, vol.~2014, no.~1, pp.~1--13, 2014.

\bibitem{tan2015clu}
W.~Tang, B.~Li, W.~Luo, and J.~Huang, ``Clustering steganographic modification
  directions for color components,'' {\em IEEE Signal Processing Letters},
  vol.~23, no.~2, pp.~197--201, 2015.

\bibitem{qin2019ano}
X.~Qin, B.~Li, S.~Tan, and J.~Zeng, ``A novel steganography for spatial color
  images based on pixel vector cost,'' {\em IEEE Access}, vol.~7,
  pp.~8834--8846, 2019.

\bibitem{wan2020non}
Y.~Wang, W.~Zhang, W.~Li, X.~Yu, and N.~Yu, ``Non-additive cost functions for
  color image steganography based on inter-channel correlations and
  differences,'' {\em IEEE Transactions on Information Forensics and Security},
  vol.~15, pp.~2081--2095, 2020.

\bibitem{tab2018ane}
T.~Taburet, L.~Filstroff, P.~Bas, and W.~Sawaya, ``An empirical study of
  steganography and steganalysis of color images in the \protect{JPEG}
  domain,'' in {\em International Workshop on Digital Watermarking},
  pp.~290--303, 2018.

\bibitem{cog2019the}
R.~Cogranne, Q.~Giboulot, and P.~Bas, ``The \protect{ALASKA} steganalysis
  challenge: A first step towards steganalysis,'' in {\em ACM Workshop on
  Information Hiding and Multimedia Security}, pp.~125--137, 2019.

\bibitem{cog2020ste}
R.~Cogranne, Q.~Giboulot, and P.~Bas, ``Steganography by minimizing statistical
  detectability: The cases of \protect{JPEG} and color images,'' in {\em ACM
  Workshop on Information Hiding and Multimedia Security}, pp.~161--167, 2020.

\bibitem{lia2016con}
X.~Liao, G.~Chen, and J.~Yin, ``Content-adaptive steganalysis for color
  images,'' {\em Security and Communication Networks}, vol.~9, no.~18,
  pp.~5756--5763, 2016.

\bibitem{abd2016col}
H.~Abdulrahman, M.~Chaumont, P.~Montesinos, and B.~Magnier, ``Color images
  steganalysis using rgb channel geometric transformation measures,'' {\em
  Security and communication networks}, vol.~9, no.~15, pp.~2945--2956, 2016.

\bibitem{gol2014ric}
M.~Goljan, J.~Fridrich, and R.~Cogranne, ``Rich model for steganalysis of color
  images,'' in {\em IEEE International Workshop on Information Forensics and
  Security}, pp.~185--190, 2014.

\bibitem{yan2020col}
C.~Yang, Y.~Kang, F.~Liu, X.~Song, J.~Wang, and X.~Luo, ``Color image
  steganalysis based on embedding change probabilities in differential
  channels,'' {\em International Journal of Distributed Sensor Networks},
  vol.~16, no.~5, p.~1550147720917826, 2020.

\bibitem{zen2019wis}
J.~Zeng, S.~Tan, G.~Liu, B.~Li, and J.~Huang, ``\protect{WISERN}et: Wider
  separate-then-reunion network for steganalysis of color images,'' {\em IEEE
  Transactions on Information Forensics and Security}, vol.~14, no.~10,
  pp.~2735--2748, 2019.

\bibitem{you2020ima}
Y.~Yousfi, J.~Butora, E.~Khvedchenya, and J.~Fridrich, ``Image\protect{N}et
  pre-trained \protect{CNN}s for \protect{JPEG} steganalysis,'' in {\em IEEE
  International Workshop on Information Forensics and Security}, pp.~1--6,
  2020.

\bibitem{chu2020ane}
K.~Chubachi, ``An ensemble model using \protect{CNN}s on different domains for
  \protect{ALASKA}2 image steganalysis,'' in {\em IEEE International Workshop
  on Information Forensics and Security}, pp.~1--6, 2020.

\bibitem{den2009ima}
J.~Deng, W.~Dong, R.~Socher, L.-J. Li, K.~Li, and L.~Fei-Fei,
  ``Image\protect{N}et: A large-scale hierarchical image database,'' in {\em
  IEEE Conference on Computer Vision and Pattern Recognition}, pp.~248--255,
  2009.

\bibitem{but2021how}
J.~Butora, Y.~Yousfi, and J.~Fridrich, ``How to pretrain for steganalysis,'' in
  {\em ACM Workshop on Information Hiding and Multimedia Security},
  pp.~143--148, 2021.

\bibitem{you2021imp}
Y.~Yousfi, J.~Butora, J.~Fridrich, and C.~Fuji~Tsang, ``Improving
  \protect{E}fficient\protect{N}et for \protect{JPEG} steganalysis,'' in {\em
  ACM Workshop on Information Hiding and Multimedia Security}, pp.~149--157,
  2021.

\bibitem{but2019rev}
J.~Butora and J.~Fridrich, ``Reverse \protect{JPEG} compatibility attack,''
  {\em IEEE Transactions on Information Forensics and Security}, vol.~15,
  pp.~1444--1454, 2019.

\bibitem{xie2017agg}
S.~Xie, R.~Girshick, P.~Doll{\'a}r, Z.~Tu, and K.~He, ``Aggregated residual
  transformations for deep neural networks,'' in {\em IEEE Conference on
  Computer Vision and Pattern Recognition}, pp.~1492--1500, 2017.

\bibitem{xu2016ens}
G.~Xu, H.-Z. Wu, and Y.~Q. Shi, ``Ensemble of \protect{CNN}s for steganalysis:
  An empirical study,'' in {\em Proceedings of the 4th ACM Workshop on
  Information Hiding and Multimedia Security}, pp.~103--107, 2016.

\bibitem{alaska}
``\protect{ALASKA} homepage.'' \url{https://alaska.utt.fr}.

\bibitem{den2015imp}
T.~Denemark and J.~Fridrich, ``Improving steganographic security by
  synchronizing the selection channel,'' in {\em ACM Workshop on Information
  Hiding and Multimedia Security}, pp.~5--14, 2015.

\bibitem{li2015str}
B.~Li, M.~Wang, X.~Li, S.~Tan, and J.~Huang, ``A strategy of clustering
  modification directions in spatial image steganography,'' {\em IEEE
  Transactions on Information Forensics and Security}, vol.~10, no.~9,
  pp.~1905--1917, 2015.

\bibitem{guo2014uni}
L.~Guo, J.~Ni, and Y.~Q. Shi, ``Uniform embedding for efficient \protect{JPEG}
  steganography,'' {\em IEEE transactions on Information Forensics and
  Security}, vol.~9, no.~5, pp.~814--825, 2014.

\bibitem{ben2009cur}
Y.~Bengio, J.~Louradour, R.~Collobert, and J.~Weston, ``Curriculum learning,''
  in {\em International Conference on Machine Learning}, pp.~41--48, 2009.

\bibitem{tan2019efficientnet}
M.~Tan and Q.~Le, ``Efficient\protect{N}et: Rethinking model scaling for
  convolutional neural networks,'' in {\em International Conference on Machine
  Learning}, pp.~6105--6114, 2019.

\bibitem{you2019bre}
Y.~Yousfi, J.~Butora, J.~Fridrich, and Q.~Giboulot, ``Breaking
  \protect{ALASKA}: Color separation for steganalysis in \protect{JPEG}
  domain,'' in {\em ACM Workshop on Information Hiding and Multimedia
  Security}, pp.~138--149, 2019.

\end{thebibliography}



\end{document}